\documentclass{article}
\usepackage{amsmath}
\usepackage{graphicx}

\usepackage{algorithm}
\usepackage{arevmath}     
\usepackage[noend]{algpseudocode}
\usepackage{float}

\usepackage{caption}
\usepackage{subcaption}
\usepackage{url}

\PassOptionsToPackage{numbers, compress}{natbib}
\usepackage[preprint]{neurips_2021}

\usepackage[utf8]{inputenc} 
\usepackage[T1]{fontenc}    
\usepackage{url}            
\usepackage{booktabs}       
\usepackage{amsfonts}       
\usepackage{nicefrac}       
\usepackage{microtype}      
\usepackage{xcolor}         

\title{Creating Powerful and Interpretable Models with Regression Networks}

\author{%
  Lachlan O'Neill\thanks{Other authors listed alphabetically} \\
  Faculty of Information Technology\\
  Monash University\\
  Clayton, Australia\\
  \texttt{lachlan.oneill@monash.edu} \\
  \And
  Simon Angus \\
  Dept. of Economics, Monash Business School \\
  SoDa Laboratories, Monash Business School \\ 
  Monash University\\
  Clayton, Australia\\
  \And
  Satya Borgohain \\
  SoDa Laboratories, Monash Business School \\
  Monash University\\
  Clayton, Australia\\
  \And
  Nader Chmait \\
  Faculty of Information Technology\\
  Monash University\\
  Clayton, Australia\\
  \And David L. Dowe \\
  Faculty of Information Technology\\
  Monash University\\
  Clayton, Australia
}

\begin{document}

\maketitle


\begin{abstract}
As the discipline has evolved, research in machine learning has been focused more and more on creating more powerful neural networks, without regard for the interpretability of these networks. Such ``black-box models'' yield state-of-the-art results, but we cannot understand why they make a particular decision or prediction. Sometimes this is acceptable, but often it is not.

We propose a novel architecture, Regression Networks, which combines the power of neural networks with the understandability of regression analysis. While some methods for combining these exist in the literature, our architecture generalizes these approaches by taking interactions into account, offering the power of a dense neural network without forsaking interpretability. We demonstrate that the models exceed the state-of-the-art performance of interpretable models on several benchmark datasets, matching the power of a dense neural network. Finally, we discuss how these techniques can be generalized to other neural architectures, such as convolutional and recurrent neural networks.
\end{abstract}


\section{Introduction}
There are many problem domains where understanding \textit{why} a particular prediction was made by a model is considered just as important as the prediction accuracy of the model, if not more important. High-stakes decision environments such as medical care, criminal justice, financial systems, and many others, require not only accurate predictions but also the rationale for why those predictions were made, for both ethical \cite{piano2020ethical} and sometimes even legal \cite{Goodman2017} reasons. Techniques for interpreting dense neural network models exist \cite{fan2021interpretability}, but they are both complex and imperfect \cite{kaur2020interpreting}. In recent years, there has been a call for further investigation into ``\textit{inherently interpretable}'' models \cite{Rudin2019}, which is undoubtedly a reaction to this eternal conflict between machine learning researchers pushing developing more complex models and practitioners trying to explain their models (especially to non-technical stakeholders).

Another solution is to use more traditional regression analysis techniques, such as simple linear/logistic regression \cite{montgomery2021introduction}, as well as variants such as Generalized Linear Models \cite{nelder1972generalized} and Generalized Additive Models \cite{hastie2017generalized}. The interpretability of these models stems from the ease at which a practitioner can quantitatively determine the effect of a particular feature on a particular prediction. However, they lack the flexibility of dense neural network (DNN) models, which have long been proven to be ``universal approximators'' \cite{hornik1989multilayer}. An ideal model would have the interpretability of a regression-based model, and the predictive power of a dense neural network model. There have been recent attempts to reconcile these two techniques \cite{agarwal2020neural}, but these models are still less powerful than true dense neural networks and do not take interactions between different variables into account.


We present a solution to this problem, which we call \textbf{Regression Networks}. These are neural networks with a specific architectural requirement - that all variables, or combinations of variables, are analyzed independently within the network, with these analyses only being combined at a final additive layer. In this sense, Regression Networks are a form of additive model, although they have the power of neural networks as well. We show the effectiveness of these Regression Networks on several benchmark classification and regression datasets, demonstrating how the architecture's ability to take interactions between variables into account leads to significantly improved performance without sacrificing interpretability and any prospects of model visualization.\footnote{In this paper, we look at several datasets from fields with the potential for significant real-world impact (whether positive or negative), such as criminal justice and medicine. The models demonstrated are early prototypes and have not been extensively tested for bias or validity. The models, figures, and statistical conclusions included in this paper are only intended for use as examples of applications of Regression Networks.}


\section{Previous Work}
As a form of Generalized Additive Model \cite{hastie2017generalized}, Regression Networks join a family of many other techniques for creating interpretable models. Other additive models include simple linear/logistic regression \cite{montgomery2021introduction} and Generalized Linear Models (GLMs) \cite{nelder1972generalized}. These models are relatively simple to understand and interpret; however, there is also much work in the literature that attempts to explain the predictions of neural networks. Fan et al. \cite{fan2021interpretability} is a recent literature review on the topic, containing over two hundred references. On the other hand, Kaur et al. \cite{kaur2020interpreting} discuss how machine learning practitioners ``over-trust'' these methods, and demonstrate how models ``designed to be inherently interpretable'', including Generalized Additive Models, are ``easier to understand'' than the DNN explanation method tested as a ``post-hoc'' alternative (SHAP \cite{lundberg2017unified}).

The most successful attempt we have found in the literature to combine the benefits of neural networks and additive models is the class of Neural Additive Models \cite{agarwal2020neural} or ``NAMs''. These NAMs learn a transformation function for each feature in a dataset, and then add the results of these functions together within a final layer. NAMs are essentially  Regression Networks with only the ``first-level'' functions, and therefore the set of Neural Additive Models can be considered as a subset of the set of Regression Models. Regression Networks provide the predictive accuracy of dense neural networks while retaining the interpretability of regression-based models. It is important to note that, while they share some similarities, our work was developed independently of the NAM paper. Nevertheless, it is important to recognize this contribution to the literature, both because it can be considered a precursor to Regression Networks, and also as it further demonstrates the desire for such approaches from researchers.

\section{Regression Networks}
A regression network takes the typical Generalized Additive Model framework \cite{hastie2017generalized}, which creates models of the form:

\begin{gather*}
y = f_1(x_1) + f_2(x_2) + ... + f_n(x_n) + \beta
\end{gather*}

Regression Networks use neural networks to learn the functions $f_1, f_2, ...$. However, they also include the further step of adding \textit{interaction functions}. For example, including all zeroth-, first-, and second-level interactions (i.e. interactions of zero, one, or two variables - note $f_\emptyset$ is just the bias term), we find:

\begin{gather*}
y = f_\emptyset + f_{(1)}(x_1) + ... + f_{(n)}(x_n) + f_{(1, 2)}(x_1, x_2) + ... + f_{(n-1, n)}(x_{n-1}, x_n)
\end{gather*}

We may choose to remove the $f_\emptyset$ term, as this bias term might otherwise include ``contributions'' from multiple features that are impossible to split up again. For our experiments, we chose to remove this bias term.

We are not limited to only learning up to second-level interactions. For $n$ features, we may learn up to any number of interactions, from $k=0$ (which would just be predicting a single value, such as the mean) to $k=n$ (which would include a function for all potential subsets of the features). This leads to large models, however, so we propose limiting $k$ to a small value (such as 2 or 3), and then including a \textit{residual function} which is simply a final neural network that attempts to learn the residual of the Regression Network (i.e. predicting the error of the network). It is not hard to see that this theoretically leads to a model no less powerful than a normal DNN. Adding this residual function into our existing model, our final model is expressed as

\begin{gather*}
y =  f_\emptyset + f_{(1)}(x_1) + \ldots + f_{(n)}(x_n) + \\
f_{(1, 2)}(x_1, x_2) + \ldots + f_{(n-1, n)}(x_{n-1}, x_n) + \ldots + \\
r(x_1, \ldots, x_n)
\end{gather*}

Currently, the model has no ``incentive'' to learn the lower-level functions. Indeed, a perfectly fine model would have all the functions output zero, except for the residual function (which acts as a normal neural network). As another example, the level-1 functions (functions that take only 1 variable as input), could be zeroed out and incorporated in the level-2 functions instead.

We can solve this problem by making an additional assertion (which is arguably an observation): \textit{In a good explanation, we would prefer the lower-level functions to explain as much as possible.} That is, all things being equal, we would prefer a model which explains itself in terms of lower-level functions than a model that explains itself using higher-level functions. We can enforce such a restriction on a model by using a ``step-wise training algorithm'':

\begin{algorithmic}[1]
\Procedure{Train}{$a,b$}
    \State Set k = 1
    \While{$k <= K_{max}$}
        \State Train $k$-level functions
        \State Fix $k$-level functions 
        \State Increment $k$ by 1
    \EndWhile
    \State Train the residual function
\EndProcedure
\end{algorithmic}

\begin{figure}
  \centering
  \includegraphics[width=\textwidth]{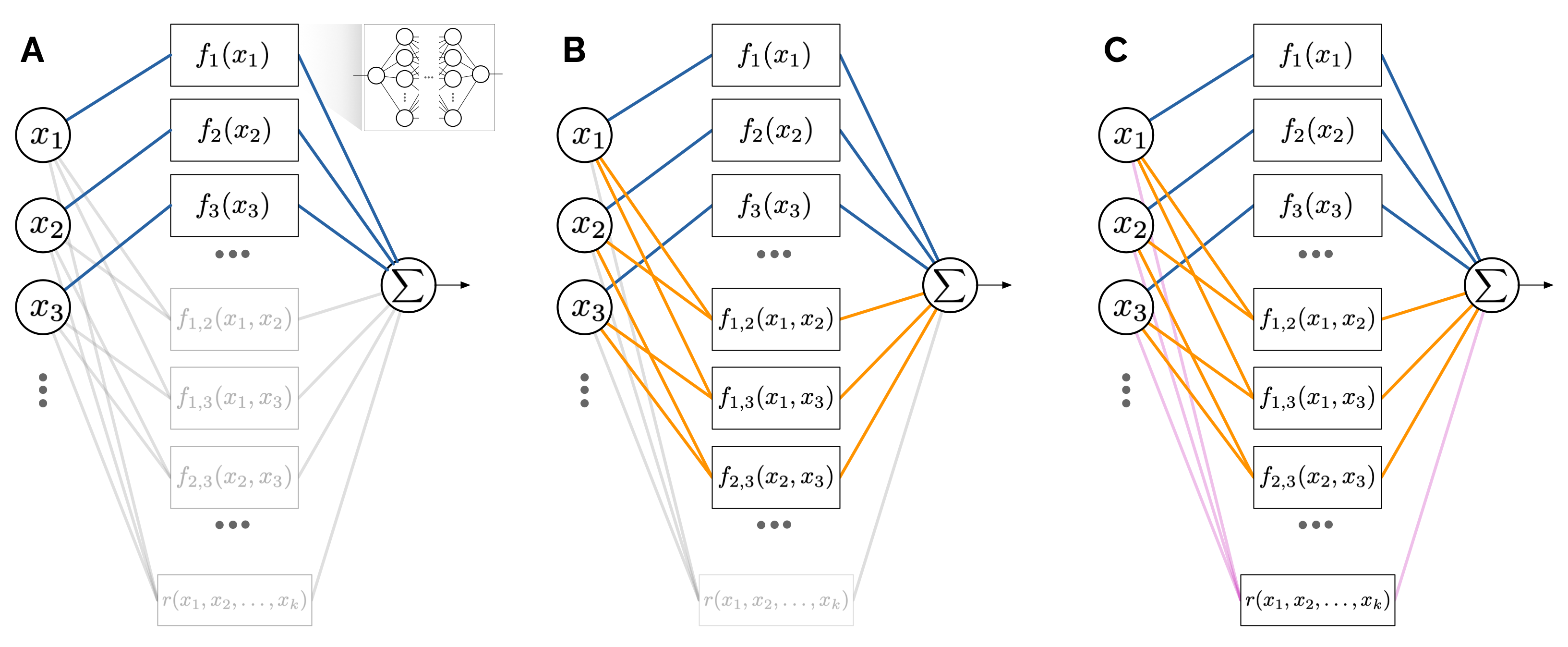}
  \caption{\textbf{Regression Network architecture and step-wise training.} In the first step, $k=1$ functions alone are learned (A); followed by $k=2$ functions (B); and finally the residual function (C).}
\end{figure}

\section{Experimental Results}
Here we compare Regression Networks to other explainable networks, as well as dense neural networks (DNNs). We demonstrate that Regression Networks are capable of performing on par with DNNs while maintaining the explainability of regression analysis. All results are reported to five significant figures. Results give the mean minimum value found on a 20\% holdout validation set on each dataset, averaged over at least three runs (with five runs for most results). The standard deviation is also included in brackets after each value. During training, class balancing was used as suggested by the TensorFlow documentation \cite{tfclassbalance}. All code used to generate the synthetic data, train the models, and evaluate them, is freely available online. All experiments were run using TensorFlow\cite{tensorflow2015-whitepaper} on Google Cloud AI Platform instances using NVidia K80 GPUs. Further information on the compute architecture, hyperparameters, dataset preprocessing, etc. can be found in the supplementary materials, as well as the aforementioned code.

All models were trained for up to 512 epochs over the respective datasets, using early stopping to monitor the validation loss and avoid overfitting (with a patience of 32 and tolerance of 0.005). No test sets were used for the experiments; as these models are not being implemented to demonstrate state-of-the-art performance on these models, the use of separate testing sets would be an unnecessary complication. All dataset manipulation operations were deterministic and used a constant random seed (except for the final data shuffling done while training), to ensure replicability of these results.

It is demonstrated by Agarwal et al. \cite{agarwal2020neural} that Neural Additive Models (which we would call first-level Regression Networks) are superior to regression \cite{montgomery2021introduction}, Generalized Linear Regression\cite{nelder1972generalized}, and Generalized Additive Models\cite{hastie2017generalized}, as well as Explainable Boosting Machines \cite{nori2019interpretml} and Gradient Boosted Trees (XGBoost) \cite{chen2016xgboost}. This is not surprising, given the inherent limitations of such approaches, and we have therefore chosen to omit these techniques from our analysis. Instead, our investigation was focused on determining the improvements yielded by adding second-level functions to the Regression Networks, as well as the residual functions. For each experiment, we demonstrate results on Linear/Logistic Regression (whichever is appropriate), Regression Networks with first-level functions only (``K1''), Regression Networks with first- and second-level functions (``K1+2''), Regression Networks with first- and second-level functions, and a residual function (``K1+2+Res''), and a final Dense model (``Dense'').

\subsection{Synthetic Datasets}
As an initial demonstration of the power of Regression Networks, as well as the importance of learning higher-level functions, we have generated several simple, artificial datasets. Variables $x$, $y$, and $q$ are numeric, and $a$ and $b$ are categorical; $f(a)$, $g(b)$, and $h(a, b)$ are mappings from these categorical values to some numerical value. For more information on distributions, etc., both the datasets and the code to generate them have been provided.\footnote{Note that a bug in the synthetic dataset generation pipeline meant that some datasets had values of $x$, $y$, and $z$ generated with $\mu=0$ and/or $\sigma=1$. We do not believe this has significantly affected the results. Regardless, all datasets have been provided in CSV form.}

\begin{table}[H]
  \caption{Synthetic Dataset Variables}
  \label{synthvariables}
  \centering
    \begin{tabular}{lll}\toprule
    Variable & Type & Distribution \\
                      \midrule
    $x$ & Continuous & $N(\mu=1, \sigma=2)$ \\
    $y$ & Continuous & $N(\mu=-4, \sigma=3.5)$ \\
    $q$ & Continuous & $N(\mu=0, \sigma=1)$ \\
    $a$ & Categorical & \{0, 1, 2, 3\} \\
    $b$ & Categorical & \{0, 1\} \\\bottomrule
    \end{tabular}
\end{table}

\begin{table}[H]
  \caption{Synthetic Dataset Descriptions}
  \label{synthvariables}
  \centering
    \begin{tabular}{ll}\toprule
    Name & Generating Function \\
                      \midrule
    Add & $z = x + y$ \\
    Add and Multiply & $z = x + y + xy$ \\
    Complex &  $z = x^2 + y - q + \sqrt{\|xz\| + 0.0001} + \log{\|y + z + 0.0001\|}$ \\
    Importance & $z = x + y + 0.2x\ln(y/100)$ \\
    Categorical & $z = f(a) + g(b)$ \\
    Categorical Interact & $z = x + y + xy + f(a) + g(b) + h(a, b)$ \\ \bottomrule
    \end{tabular}
\end{table}

\begin{table}[H]
    \caption{Synthetic Dataset Performance (Mean Squared Error)}
    \label{Synthetic}
    \centering
    \begin{tabular}{@{}llllll@{}}
    \toprule
    \textbf{Dataset}               & \multicolumn{1}{c}{\textbf{Regression}} & \multicolumn{1}{c}{\textbf{K1}} & \multicolumn{1}{c}{\textbf{K1+2}} & \multicolumn{1}{c}{\textbf{K1+2+Res}} & \multicolumn{1}{c}{\textbf{Dense}} \\ \midrule
    \textbf{Add}          & \textbf{0 (0)}                            & \textbf{0 (0)}                  & \textbf{0 (0)}                    & \textbf{0 (0)}                        & \textbf{0 (0)}                     \\
    \textbf{Add \& Multiply}                 & 47.476 (0.4)                            & 37.9097 (18.96)                 & 0.0055 (0)                        & \textbf{0.0036 (0)}                   & 0.0055 (0)                         \\
    \textbf{Complex}        & 61.6533 (0.07)                          & 61.877 (0.21)                   & \textbf{0.0043 (0)}               & 0.0072 (0)                            & 0.0058 (0)                         \\
    \textbf{Importance}     & 0.0004 (0)                              & 0.0005 (0)                      & \textbf{0 (0)}                    & 0.0003 (0)                            & 0.0002 (0)                         \\
    \textbf{Categorical}        & 0 (0)                                   & 0 (0)                           & 0 (0)                             & 0 (0)                                 & \textbf{0 (0)}                     \\
    \textbf{Categorical Interact}       & 97.2999 (0.02)                          & 97.3668 (0.08)                  & \textbf{0 (0)}                    & \textbf{0 (0)}                        & \textbf{0 (0)}                     \\ \bottomrule
    \end{tabular}
\end{table}

These datasets have been implemented to validate both the codebase used for these experiments and the underlying theoretical understanding of Regression Network models. In particular, note that the K2 and Dense models fully converge on all datasets. The Regression model can only fully converge on linear functions (Add and Categorical). The K1 models converge on the same models (and perform between regression and K2 models otherwise, which is the expected result). All of these results are expected due to the limitations of each model and the requirements of each dataset.

\subsection{Real Datasets}

\subsubsection{Regression Problems}
We used Regression Networks to analyze three regression datasets: the ProPublica COMPAS dataset (``COMPAS'') \cite{dressel2018accuracy}, the Boston Housing Market dataset (``Boston'') \cite{bostondata}, and the Californian Housing Market Dataset (``California'') \cite{calidata}. These proved to be useful tests, containing selections of both numerical and categorical variables.

\begin{table}[H]
  \caption{Regression Problem Performance (Mean Squared Error)}
  \label{regression}
  \centering
    \begin{tabular}{@{}llllll@{}}
    \toprule
    \textbf{Dataset}               & \multicolumn{1}{c}{\textbf{Regression}} & \multicolumn{1}{c}{\textbf{K1}} & \multicolumn{1}{c}{\textbf{K1+2}} & \multicolumn{1}{c}{\textbf{K1+2+Res}} & \multicolumn{1}{c}{\textbf{Dense}} \\ \midrule
    \textbf{COMPAS \cite{dressel2018accuracy}}                & 0.7932 (0)                              & 0.7594 (0)                      & \textbf{0.7028 (0)}               & 0.7086 (0.01)                         & 0.7063 (0.01)                      \\
    \textbf{Boston \cite{bostondata}}       & 0.1894 (0.02)                           & 0.0981 (0)                      & 0.0843 (0.01)                     & 0.0886 (0.03)                         & \textbf{0.0468 (0)}                \\
    \textbf{California \cite{calidata}}   & 0.3767 (0)                              & 0.3086 (0.03)                   & \textbf{0.2079 (0)}               & 0.2382 (0.07)                         & 0.2346 (0)                         \\ \bottomrule
    \end{tabular}
\end{table}

On the COMPAS and California datasets, the K2 model was competitive with the Dense model. While it did lose out on the Boston dataset, it still gives significantly better performance than regression and the K1 model.

\subsubsection{Classification Problems}
Three classification benchmarks from the UCI dataset repository \cite{ucirepository} were used to test the effectiveness of Regression Networks on classification problems: Census Income (commonly known as ``Adult'') \cite{adultdata}, Diabetes Readmission (``Diabetes'') \cite{diabetesdata}, and Mammographic Mass (``Mammogram'').

\begin{table}[H]
    \caption{Classification Problem Performance (Binary Cross-Entropy)}
    \label{classification}
    \centering
    \begin{tabular}{@{}llllll@{}}
    \toprule
    \textbf{Dataset}               & \multicolumn{1}{c}{\textbf{Regression}} & \multicolumn{1}{c}{\textbf{K1}} & \multicolumn{1}{c}{\textbf{K1+2}} & \multicolumn{1}{c}{\textbf{K1+2+Res}} & \multicolumn{1}{c}{\textbf{Dense}} \\ \midrule
    \textbf{Adult \cite{adultdata}}                 & 0.4334   (0)                            & 0.3625 (0)                      & \textbf{0.3395 (0.01)}            & 0.3772 (0.01)                         & 0.3715 (0.01)                      \\
    \textbf{Diabetes \cite{diabetesdata}}              & 0.4837 (0.03)                           & 0.583 (0.02)                    & \textbf{0.4125 (0.01)}            & 0.53 (0.03)                           & 0.551 (0.02)                       \\
    \textbf{Mammogram \cite{mammogramdata}}             & 0.5037 (0)                              & \textbf{0.493 (0)}              & 0.5093 (0.01)                     & 0.5067 (0.01)                         & 0.4948 (0.01)                      \\ \bottomrule
    \end{tabular}
\end{table}

As with the regression problems, the results demonstrate the power of the Regression Network architecture. In the Adult and Diabetes datasets, the addition of second-level functions significantly increases the power of the model compared to regression or first-level function models. The Mammogram dataset shows similar performance across all five architectures, with the first-level Regression Network model taking the win on that benchmark.

\subsection{Interaction Plots}
One of the key benefits of Regression Networks is that it is possible to visualize not just the single variable effects, but also the effects of combinations of multiple variables. This section will contain a brief overview of several selected plots. Many more are available in the supplementary materials.

\begin{figure}[H]
  \centering
  \includegraphics[width=\textwidth]{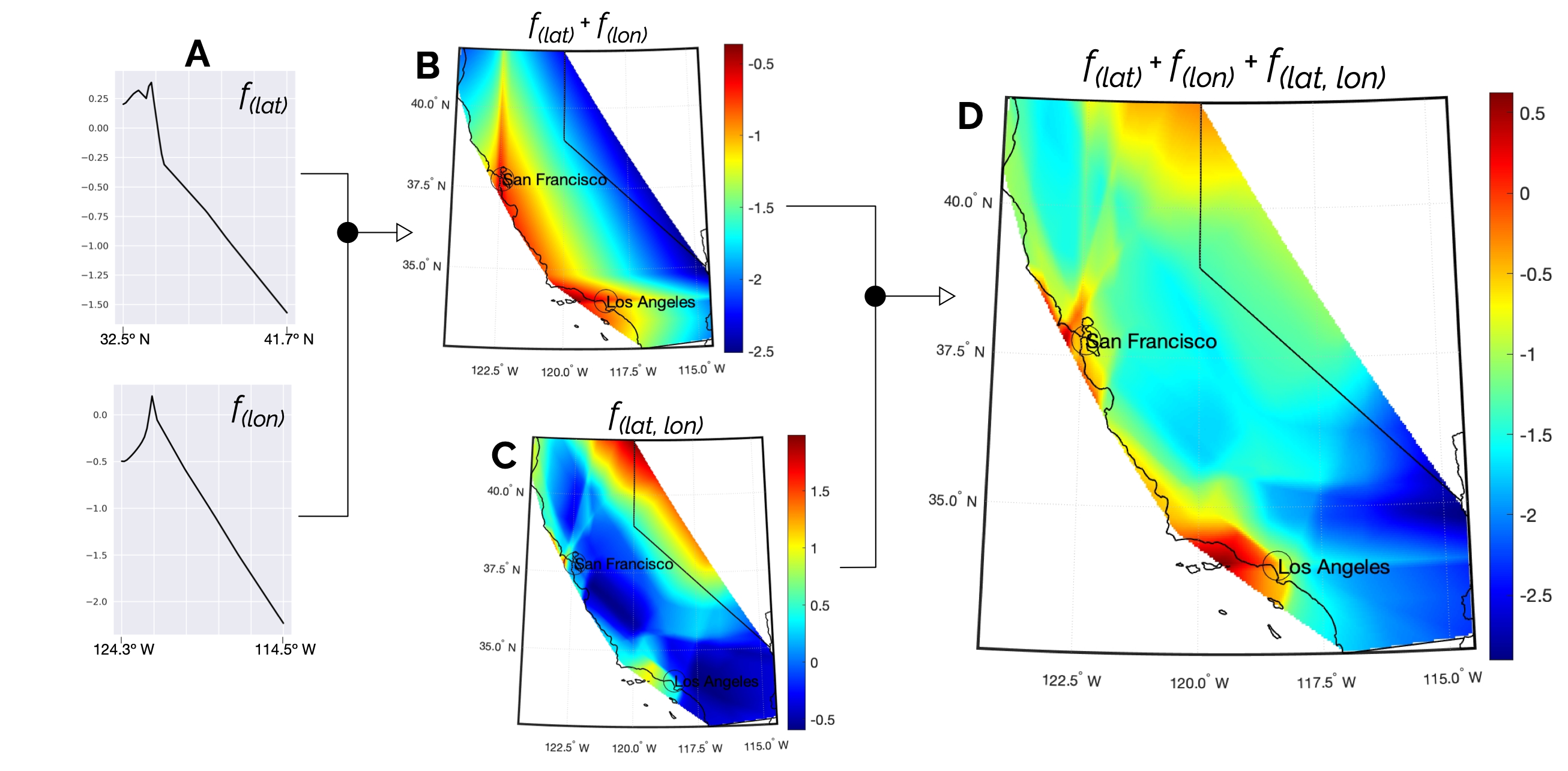}
    \caption{\textbf{Regression Networks learning Latitude and Longitude functions in the Californian Housing dataset.} In the step-wise method, $k=1$ level functions are learned (A); these are combined and overlaid on geographic data (B); then, the interaction $k=2$ function is learned (C); and finally, the combined $k=1$ and $k=2$ outcome is obtained (D).}
\end{figure}

\begin{figure}[H]
  \centering
  \includegraphics[width=\textwidth]{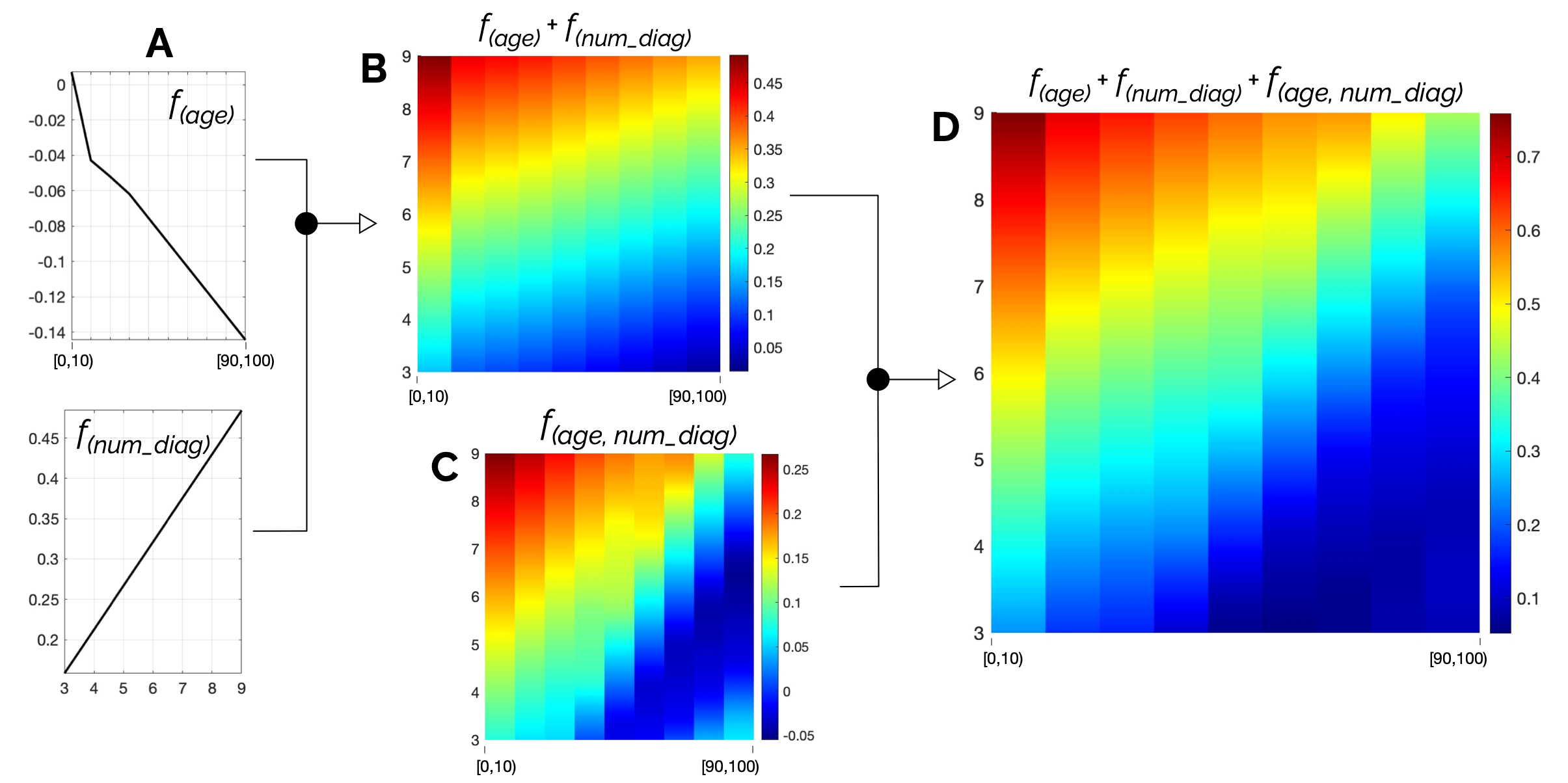}
    \caption{\textbf{Regression Networks learning Age and Diagnosed Illnesses functions in the Diabetes Readmission dataset.} In the step-wise method, $k=1$ level functions are learned (A); these are combined (B); then, the interaction $k=2$ function is learned (C); and finally, the combined $k=1$ and $k=2$ outcome is obtained (D).\textbf{This is not a medical experiment, and these results are experimental and not intended for real-world use.}}
\end{figure}

Consider the plot in Figure 2 of California overlaying the interaction function between ``latitude'' and ``longitude'', trained on the California Housing dataset. Note the hot spots in San Francisco and Los Angeles, which correspond to higher housing prices in these locations (which is not surprising). Also notice how San Francisco and Los Angeles are selected as individual regions, rather than a particular latitude/longitude (as seen in the 2D plot including only the first-level functions). Doing this accurately would not be as successful without the inclusion of interactions, as the best the model can do is determine an average effect at a particular latitude \textit{or} longitude - not both at once.

It is important to note that the model has not simply learned the \textit{existence} of a correlation between a certain location in California (as a latitude and longitude pairs) and an increase in housing price; that is not a difficult conclusion to reach with any reasonable model. The advantage of this approach is that it also numerically shows the unique effect of this correlation, distinct from all other features (although correlated features might mean that some portion of the effect of one feature is included in the functions of other variables instead - unfortunately, this is unavoidable when features are correlated).

Figure 3 demonstrates a different type of insight second-level Regression Networks can provide. The shape of Figure 3.B (containing only first-level functions) and 4.D (containing first-level functions with second-level functions added) look similar and lead to the same qualitative conclusion: that as people get older, a higher number of diagnosed illnesses is required to contribute the same amount to the probability of requiring hospital readmission. However, notice how the maximum value in Figure 3.B is approximately 0.5, while for Figure 3.D it is approximately 0.75. The quantitative insight gleaned from the functions generated by Regression Networks, in this case, indicates that the impact of the number of diagnoses in young people might be stronger than the first-level functions show. The positive contribution to the logit of requiring readmission increases by 50\% when taking the interaction function into account. \footnote{Note that these are simply the observations one can make from the Regression Network model trained on this dataset - we do not claim any medical validity, and this is not intended for medical use.}




\subsection{Training Without the Step-Wise Algorithm}
A reasonable question to ask is whether the step-wise training algorithm is truly necessary to obtain good results. It might be the case that, in real-world problems, it is sufficient to train the entire model at once. We have evaluated this experimentally and determined that it is indeed necessary to use the step-wise training algorithm, as the model may not successfully learn lower-level functions without it. Discussion of this, including exemplary plots, is included in the supplementary materials.

\section{Discussion}
The experimental results demonstrate the superior performance of the second-level Regression Networks, compared to simple regression and Neural Additive Models (or NAMs \cite{agarwal2020neural}, which are first-level Regression Networks). The results for the second-level networks were comparable to the corresponding DNNs, most often meeting or exceeding them in performance. However, we do note that more time was naturally spent experimenting with hyperparameters and the like while developing the Regression Network codebase, and it is likely that the DNNs could have improved performance with proper hyperparameter tuning (although this is also the case for the Regression Networks). Due to computational limitations, we omitted this hyperparameter tuning step and instead chose to fix the set of hyperparameters used by each algorithm over all experiments. Importantly, we do not claim that the Regression Networks are inherently capable of superior performance to DNNs - our theoretical understanding indicates that at best they should meet the performance of a DNN.

\subsection{Value of the Residual Function}
The limited utility of the addition and training of the residual function was a surprising result. This will require further experimentation; however, with careful implementation and hyperparameter tuning, the addition of the residual function is possibly still useful in some cases.

\subsection{Combinatorial Explosion}
One issue with the architecture is that a large number of features can lead to a combinatorial explosion as the number of functions learned in the second phase of training grows with the square of the number of features. In particular, one-hot encoded variables which encode categorical variables with a large number of potential values can cause the number of functions in the model to explode very quickly; this problem compounds when multiple such variables exist in the dataset. There are several potential solutions to this particular problem. In our work, we simply disabled one-hot encoding for the two affected datasets: Adult and Diabetes. This disadvantaged the regression models, but since our main focus is on the comparison between the Regression Networks and DNNs, we found this acceptable.

This approach was successful with the datasets used in our experiments; however, there are potentially better solutions. For example, variable pruning could be used to determine which one-hot values are important. Alternatively, a preprocessing ``clustering'' step could be used to combine the one-hot values into several distinct sets and use that instead of the original values. Finally, there are optimizations discussed in the supplementary materials that help alleviate this problem somewhat.

\subsection{Learning from Correlated Features}
One concern with Regression Networks is the way they deal with correlated features. For example, suppose a first-level Regression Network is being trained on a simple dataset $z = x + y$, where $x = 2y + 3$. In this case, there are arbitrary many pairs of functions that would solve the problem exactly. It might well learn the functions $f_x(x) = 0$ and $f_y(y) = 3y + 3$ - this would give the same accuracy as the intended functions $f_x(x) = x$ and $f_y(y) = y$. This problem is not unique to Regression Networks, and correlated features are a source of concern within the field of data science \cite{hall1999correlation}. It is therefore very important with Regression Networks, as it is with all machine learning models, to ensure that features are as little correlated with one another as possible. In particular, it is not useful to include transformations of a particular variable in the analysis, both because they cause the two variables to become correlated, and also because the Regression Network is already capable of transforming any variables by an arbitrary transformation function.

\subsection{Datasets With Overwhelming High-Level Interactions}
In the dataset formed by the function $z = x + y + xy$, the term $xy$ overwhelms the impact of the terms $x$ and $y$ because, for $\|x\|, \|y\| > 1$, $\|xy\| > \|x\| + \|y\|$. In this case, the step-wise training of the first-level functions may struggle to learn appropriate functions, and produce nonsensical functions instead. This is not a particular failing of the model, but rather due to the unavoidable fact that such a function simply cannot be successfully approximated by a sum of terms containing only $x$ or $y$. Plots demonstrating this problem are included with the supplementary materials. However, we conjecture that in most real-world problems it is unlikely that interactions between variables will be significantly more important than the individual variables themselves (although it will probably happen sometimes). If these interaction terms are overwhelming the lower-level terms and leading to nonsensical functions, we suggest trying the ``all-in-one'' training approach (training all functions at once) rather than using the step-wise algorithm.

\subsection{Classification with Three or More Classes}
Binary classification problems are trivial to implement using Regression Networks, as the regression output can be simply considered as a logit from which a probability for each class can be calculated. However, this does not generalize to problems with three or more classes, and another approach is necessary. We propose a simple ``one-vs-rest'' approach, where a Regression Network model is trained, for each class, on the binary classification problem of belonging to that class or any other class. This technique is also used in Logistic Regression to solve this problem \cite{sejnowski1987parallel}, so it is a reasonable choice. This will be explored in further research.

\section{Further Work}
We are developing several potential extensions of Regression Networks that have the potential to extend the benefits of this architecture to these more complex problem domains. The main challenge is identifying understandable features which can be separately fed into the model. Some preliminary commentary on this is provided in the supplementary materials. For example, consider a simple image classification problem (such as classifying clothing as in the ``MNIST Fashion'' dataset \cite{xiao2017fashion}). We propose an architecture which learns a set of $N$ ``orthogonal functions'' on an input image $X$, i.e. $f_1(X), f_2(X), ..., f_N(X)$ such that, for any pair $1 <= a, b <= n, a \neq b$, $f_a(f_b(x)) = 0$. Essentially, this is splitting the image up into separate components. We conjecture that using such transformations, and passing them in as features to a Regression Network with Convolutional Neural Networks as the function approximators for each feature, would provide a successful, interpretable Regression Network model.

Another current limitation of the current approach is that it only provides point estimates. One potential solution to this problem is to have each function in the Regression Network output a distribution as the predicted effect of a particular variable on the output variable. Assuming a Gaussian distribution is the output of each function, we can then find a distribution for the output variable by using the trivial methods for finding the distribution of a sum of Gaussian distributions.

Finally, we intend to investigate potential methods for co-training of different function levels. For example, it might be good to allow small changes to the first-level functions while training the second-level functions. This is discussed in the supplementary materials. One potential method for this is to use different learning rates for the different function levels, so the first-level functions are more resistant to change than the second-level functions. This can of course be generalized to the training of any function level. While this does more hyperparameters to tune (the learning-rate for each level's functions), we conjecture that this approach could lead to models which express the true impact of each function level better than either the step-wise approach.

\section{Broader Impact}
Machine learning and AI-based decision making are already being evaluated and/or used for important decision making in fields such as medicine \cite{diabetesdata} and criminal justice \cite{fan2021interpretability}. While any technology has the capacity for both positive and negative use cases, it is our strong belief that improving the accuracy and interpretability of machine learning models will lead to net-positive results for society. It is first the responsibility of people using these models to ensure they are used only for good - and where that fails, it is the responsibility of society to seek out and put an end to situations where these models are used for bad or even malicious reasons.



\bibliographystyle{unsrt}
\bibliography{bib}

\end{document}